\documentclass[letterpaper, 10 pt, conference]{ieeeconf} 
\usepackage{times}
\usepackage{amsfonts}
\usepackage{amsmath}
\usepackage{graphicx}
\usepackage{subcaption}
\graphicspath{ {imgs/} }
\usepackage{tabularx,booktabs}

\newcolumntype{C}{>{\centering\arraybackslash}X} % centered version of "X" type
\newcommand\given[1][]{\:#1\vert\:}
\newcommand{\norm}[1]{\left\lVert#1\right\rVert}

\IEEEoverridecommandlockouts 
\title{\LARGE \bf
An Accelerated Approach to Safely and Efficiently \\ Test Pre-Production  Autonomous Vehicles on Public Streets
}

\author{Mansur Arief$^{1}$, Peter Glynn$^{2}$, and Ding Zhao$^{1}$
\thanks{* Corresponding author: Ding Zhao Email: ({\tt\small dingzhao@cmu.edu})}%
\thanks{$^{1}$Mansur Arief and Ding Zhao are with Carnegie Mellon University, USA.}%
\thanks{$^{2}$Peter Glynn is with Stanford University, USA.}%
}

\begin{document}

\maketitle
\thispagestyle{empty}
\pagestyle{empty}

%%%%%%%%%%%%%%%%%%%%%%%%%%%%%%%%%%%%%%%%%%%%%%%%%%%%%%%%%%%%%%%%%%%%%%%%%%%%%%%%
\begin{abstract}

Various automobile and mobility companies, for instance Ford, Uber and Waymo, are currently testing their pre-produced autonomous vehicle (AV) fleets on the public roads. However, due to rareness of the safety-critical cases and, effectively, unlimited number of possible traffic scenarios, these on-road testing efforts have  been acknowledged as tedious, costly, and risky. In this study, we propose Accelerated Deployment framework to safely and efficiently estimate the AVs performance on public streets. We showed that by appropriately addressing the gradual accuracy improvement and adaptively selecting meaningful and safe environment under which the AV is deployed, the proposed framework yield to highly accurate estimation with much faster evaluation time, and more importantly, lower deployment risk. Our findings provide an answer to the currently heated and active discussions on how to properly test AV performance on public roads so as to achieve safe, efficient, and statistically-reliable testing framework for AV technologies.

\end{abstract}

%%%%%%%%%%%%%%%%%%%%%%%%%%%%%%%%%%%%%%%%%%%%%%%%%%%%%%%%%%%%%%%%%%%%%%%%%%%%%%%%
\section{INTRODUCTION}

Many of the industrial leaders in automobiles, mobility providers, and intelligent systems in general are currently deploying their autonomous vehicle (AV) fleets on the public streets to learn from the real world traffic \cite{fagnant2015preparing}. While the recent advances in robotics, machine learning, and computation techniques have enabled these fleets to dynamically upgrade its perception, prediction, and control capabilities, the vast scenario space as well as the high degree of uncertainties of the traffic systems have rendered these AV deployment efforts to be overly cautious, extremely costly and time-consuming. 

The predominant reason for this is the low exposure of the safety-critical events, such as crashes, that makes the training dataset sufficient for making meaningful statistical inference to be extremely large to the extent that collecting it under Naturalistic Field-Operational Test (N-FOT), i.e.  on-road deployment, becomes time-consuming, risky, and costly \cite{wang2017much, zhao2016accelerated, greene2011efficient}. Google cars, Waymo, for instance, have logged over 3.5 million miles from four states in the USA: Washington, California, Arizona, and Texas \cite{waymo2017}. While may seem large, this million-miles driving dataset provides only  a few observations of safety-critical events that can be used for evaluation purposes. Thus, obtaining precise statistical estimates for the safety performance with high confidence level requires even much larger data. \cite{kalra2016driving}, for instance, estimated an astronomical amount of driving miles required to obtain proper estimates for AV driving, which is simply unaffordable especially for those in the pre-production phases.

The fatal accident involving an AV failing to respond to a pedestrian at night in Tempe Arizona, as well as several other reported AV failures, should effectively remind us that although such adverse events are rarely observed relative to the logged driving miles, once happened, the impact is catastrophic \cite{stilgoe2018,calDMV}. Furthermore, the inherent nature of the AV intelligence exacerbates the effectiveness of conventional predefined scenario testing approaches such as test-matrix,  or worst-case evaluation. The main reason is that these methods might fail to assess the `true' capability of the AV system as it could be designed to excel `only' under these predefined test scenarios. Meanwhile, if one were to assess its worst-case performance, information related to the probability of occurrences of such worst-case scenarios in real world is omitted and hence, an overly-conservative evaluation result will likely be concluded.

Addressing the above challenges, our contribution is predominantly to develop a safe and efficient way to test pre-produced AVs on public roads. The `pre-produced' term here implies that the capability of AVs in question is yet to be assessed. We use the driving dataset from the Safety Pilot Model Deployment (SPMD) to study the proposed framework. In addition, a robust stochastic model for traffic system and various recent advances in machine learning and adaptive design theories to construct a surrogate model for the capability are leveraged upon. As constructed, the accuracy of the surrogate model will gradually increase as more data are observed from the real-world deployments. With this improved estimation accuracy, we could purposefully select a scenario to evaluate at each deployment iteration based on its estimated risk and potential learning gain. It has been shown, at least through numerical experiment, that with such framework, fewer number of AV deployments would be needed to achieve the same accuracy level as that from random deployment (see for instance \cite{huang2017sequential} in addition to our experiment). Emphasizing on this notion of acceleration, we will refer to the proposed framework as 'Accelerated Deployment' hereafter. It is worth noting that Accelerated Deployment is applicable not only to AV test but also adaptable to the evaluation of intelligent system facing rare but safety-critical events such as field robots, medical assistive technologies, smart cities, etc. 

The rest of this work will be presented as follows. In Section II we will concisely summarize the idea constituting the Accelerated Deployment framework. In Section III we will formulate the AV deployment problems in details. In Section IV we provide numerical experiment using the publicly available data and present our initial findings and discussions about the potentials and plausible shortcomings. Finally, we conclude and provide our future directions in Section V.

\begin{figure*}[h]
\centerline{
\includegraphics[width=0.7\linewidth]{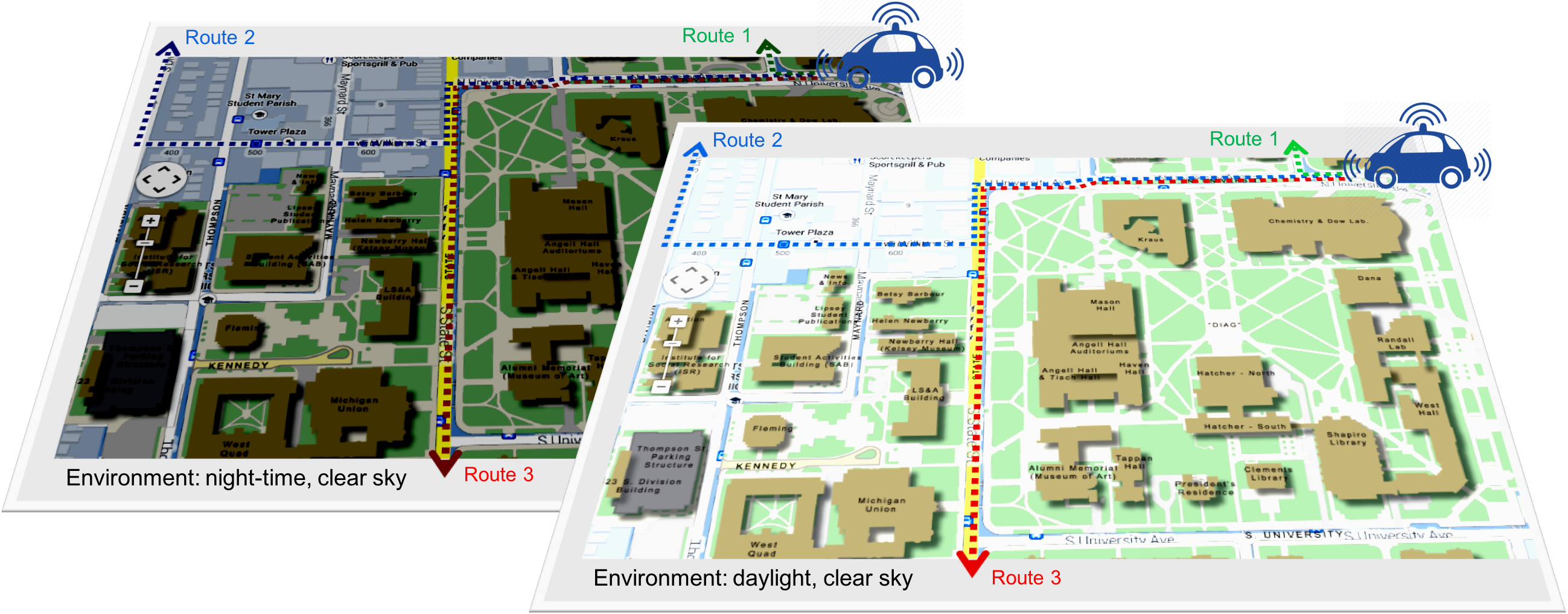}
}
\caption{AV deployment problem deals with selecting environment to deploy a pre-produced AV on public streets}
\label{framework}
\end{figure*}

\section{Accelerated Deployment Framework}
The underlying idea of Accelerated Deployment framework is to strategically pick an environment to test the AV fleet. In simpler case, one can choose to evaluate the AVs in daytime on a particular route or other alternatives routes. Or perhaps even at night-time on some other routes (see Figure \ref{framework} for the overall framework illustration). Similar idea has been proposed under design of experiment framework \cite{cox2017design}. For our case, various aspects including safety risk, potential learning, deployment cost, etc. can be used as determining factor when picking the environment. In contrast to random deployment that tests the AV under randomly picked environment, the selective picking scheme in the Accelerated Deployment framework raises concerns about the completeness of the approach in exploring the environment space. In other words, one has to be aware that such selectivity violates the randomness assumption that is inherent in the random deployment, and thus, an unaware inference based on these selectively picked environments might be biased.

This particular type of bias manifests itself in the form of `local confidence', meaning that the test results might not be generalizable to the whole environment space. Ignoring this notion of `local confidence' could yield to prematurely terminating the deployment iterations and over-confidently declaring the statistical confidence of the test results. In an iterative procedure such as our deployment problem, a proper terminating criteria is key to overcome the bias. In this study, we use multiple terminating criteria to fulfill this goal: (i) the estimation error at the final iteration should be at most some predefined tolerance $\tau$ to ensure the accuracy at the final iteration and (ii) the minimum number of deployment $\bar{N}$ should be sufficiently large so as to avoid terminating prematurely. However, we make a careful note that various termination criteria design can be embedded to improve the proposed method.

It is worth mentioning that problems dealing with making an inference from biased samples or experiments has been extensively studied, especially in the context of rare event analysis. One such interesting approach pertaining to AV evaluation studies is the Accelerated Evaluation (AE) framework \cite{zhao2016accelerated, zhao2017accelerated}. The main idea of AE is to `skew' the sampling distribution from which an evaluating scenario is drawn so that the frequencies of observing a failure is amplified. Next, importance sampling mechanism \cite{glynn1989importance} is deployed to `skew' the test result back when making a real-world inference. It has been shown that when the skewing distribution is optimal, the testing duration can be reduced to several orders of magnitude while maintaining statistical accuracy.

\section{Problem Formulation}
Consider a deployment problem for an AV with static but unknown capability prior to the evaluation. Let the capability be measured as the probability of encountering risky events $f(E)$ when the AV is deployed under environment $E \in \mathcal{E}$. We note that other types of capability measures can also be used. In our case, $f$ is a mapping from $d$-dimensional space to one-dimensional closed subspace of real number $[0,1]$. We note that $E$ could be any multivariate variable representing various information related to the environment under which an AV fleet is deployed. Such information could be traffic situations, lane characteristics, weather conditions, or other environmental features that affect the capability of the AVs. As a surrogate model for $f$, we have $\hat{f}_\theta$ with $\theta \in \Theta$ represent the parameters for $\hat{f}$. Assume that with some optimal parameter $\theta^*$, $f \approx \hat{f}_\theta^*$. Our goal, then, is to obtain $\theta^*$ with which we could accurately estimate the capability of the AVs under any given $E$ as $\hat{f}_\theta^*(E)$.

Suppose that one has collected $n$ pairs of deployment environments and the corresponding real-world observed capabilities $ \left \{ \tilde{E}^n, f\left (\tilde{E}^n \right) \right \}$, where $$\tilde{E}^n = \left[ E^{(1)}, E^{(2)}, \cdots, E^{(n)} \right]^T$$ and $$f \left(\tilde{E}^n \right) = \left[ f \left(E^{(1)} \right), f \left(E^{(2)} \right), \cdots, f \left(E^{(n)} \right)\right]^T.$$ Each $E^{(i)}$ represents the deployed environment at iteration $i$, $i=1,2,\cdots,n$. With this dataset, one can get an approximation for the parameters given the dataset $\hat{\theta} \given[\Big] \tilde{E}^n, f\left (\tilde{E}^n \right)$ by solving
\begin{align}
	z^{(n)} = \displaystyle \underset {\theta \in \Theta} {\text{min }} \norm {f\left (\tilde{E}^n \right) - \hat{f}_\theta \left (\tilde{E}^n \right)},
    \label{theta_hat}
\end{align}
where the operator $\norm{\cdot}$ returns the norm of the inside term. $\hat{\theta} \given[\Big] \tilde{E}^n, f\left (\tilde{E}^n \right)$, which will be denoted as $\theta^{(n)}$ hereafter, is the minimizer of (\ref{theta_hat}). One shall notice that by combining $f \approx \hat{f}_\theta^*$ with (\ref{theta_hat}),  it follows that as $n \rightarrow \infty, \theta^{(n)} \rightarrow \theta^*$. This implication is essentially what assures that random deployment framework would asymptotically yield to accurate capability estimation despite of the high risk and large number of iterations it might be associated with.

Suppose now that we are to select the $n+1$-th deployment scenario $E^{(n+1)}$. We can view $z(n)=z^{(n)}/n$ as the average estimation uncertainty associated with the $n$-th iteration parameter $\theta^{(n)}$. Accounting for this uncertainty, which is critical especially when $z(n)$ is `noticeably large' relative to $f\left(\tilde{E}^n \right)$ values,  we define a nondecreasing function $\alpha (\cdot)$ in such a way that $\alpha (n)$ converges to 1 as $z(n) \rightarrow 0$. The sequence of $\alpha$ values for $n \in \mathbb{N}$, denoted hereby as $\alpha^{(n)}$ will then be used as the `weight' describing the averseness of the decision makers towards the estimation uncertainty $z(n)$. 

Let $g(E)$ denote learning gained from deployment under $E$. We use estimation error associated with $E$ as a proxy to quantify $g(E)$. As such, in the case we have observed $E$, we will use the difference between our current best estimate with the real-world value. Otherwise, we use the difference between two consecutive best estimates, $\hat{f}_{\theta^{(n)}}(E)$ and $\hat{f}_{\theta^{(n-1)}}(E)$.
\begin{align}
g(E)=
\begin{cases}
f(E)-\hat{f}_{\theta^{(n)}}(E), & E \in \tilde{E}^n\\
\hat{f}_{\theta^{(n)}}(E) - \hat{f}_{\theta^{(n-1)}}(E), & \textrm{otherwise}
\end{cases}
\label{learngain}
\end{align}
With (\ref{learngain}), we obtain $E^{(n+1)}$ as the maximizer of
\begin{align}
\displaystyle \underset {E \in \mathcal{E}^{(n+1)}} {\text{max }} g(E),
\label{findingE}
\end{align}
where 
\begin{align}
\mathcal{E}^{(n+1)} = \left \{ \bar{E}: \alpha^{(n)} \hat{f}_{\theta^{(n)}} \left( \bar{E} \right) + \left( 1-\alpha^{(n)} \right) z(n)  \leq \xi \right \}
\label{feasibleE}
\end{align}
is the feasible region for the $n+1$-th deployment environment. Given some fixed risk tolerance $\xi$, this feasible region will expand as $n$ increases because the term $\left(1-\alpha^{(n)} \right) z(n)$ decreases in $n$ by construction. Such an expansion of the feasible region allows (\ref{findingE}) to return riskier environment $E^{(n+1)}$ as the estimation accuracy improves in the number of deployments $n$.

In the following subsections, we will delve into how we define the probability of engaging in risky events and the form of the surrogate model used in regard to this formulation.

\subsection{Probability of Engaging in Risky Events}
The notion of capability of an AV system can be seen as how safe it can interact with other road users under various deployment situations. If we treat its capability as static, then the various deployment situations can well be described by the various environments under which the AV system is deployed. In other words, we can view it as `how well the AV system can avoid engaging in risky events under the various environments it is deployed'. Given the structure of the available driving database that provides the state of the vehicle and its distance and relative speed to other encountered road users, the later view seems more suitable in quantifying the AV system capability. Moreover, this approach has also been extensively used by other researchers in quantifying the risk related to driving behavior or safety risk \cite{leblanc2013longitudinal, wang2018extracting}. More sophisticated capability models are also available, for instance those accounting for the foundational driving behavior and encounters \cite{wang2018understanding, laugier2011probabilistic}, but we employ the simpler ones in this study to focus on the deployment problem itself.

\begin{figure}
\centerline{
\includegraphics[width=\linewidth]{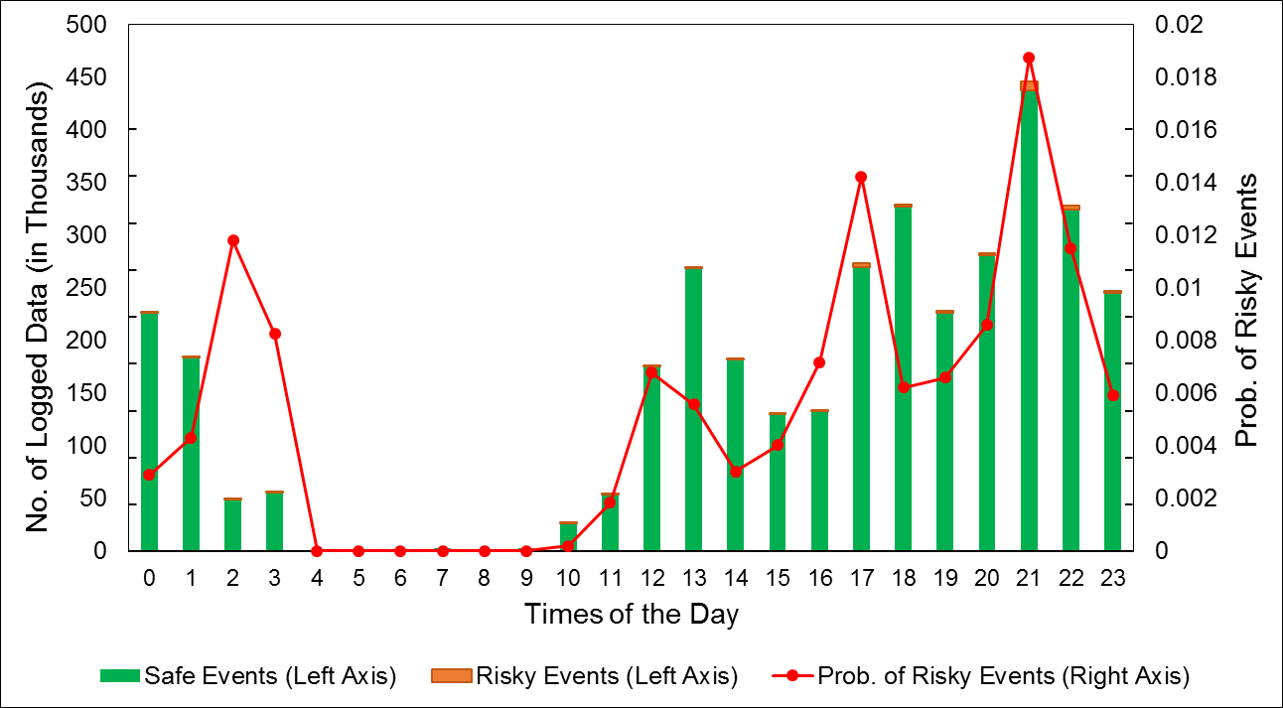}
}
\caption{The temporal trend of probability of risky events over a 24-hour period}
\label{risk_time}
\end{figure}

\begin{figure}
\centerline{
\includegraphics[width=\linewidth]{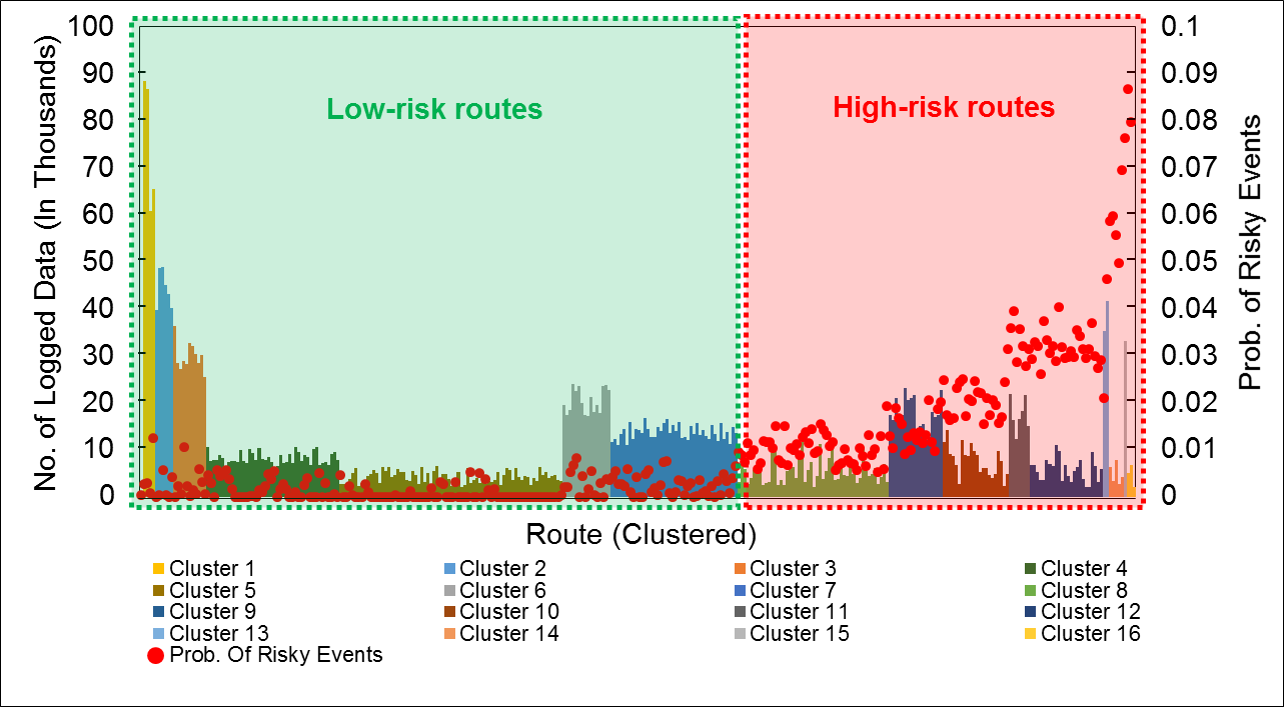}
}
\caption{The spatial trend of the probability of risky events over the routes (the 329 routes are pre-segmented into 16 route clusters) }
\label{risk_space}
\end{figure}
Suppose that we have some criteria for the risky events of interest $\varepsilon$. An event $x$ is said to be risky if $\mathbb{I}(x \in \varepsilon)=1$. The function $\mathbb{I}(\cdot)$ is an indicator function returning value 1 if the argument is true, and 0 otherwise. Thus, the probability of encountering risky events under environment $E$ is
\begin{align}
\mathbb{P}(E) = \mathbb{E}_x[\mathbb{I}(x \in \varepsilon)].
\label{expect}
\end{align}
If we have sufficiently large $N$ instances of the random events, i.e. \{$x^{(1)}, x^{(2)}, \cdots, x^{(N)}$\}, then 
\begin{align}
\hat{p}(E) = \frac{1}{N} \sum_{j=1}^N {\mathbb{I} \left( x^{(j)} \in \varepsilon \right)}
\label{expind}
\end{align}
provides unbiased estimator for (\ref{expect}). Thus, we use $\hat{p}(E)$ in (\ref{expind}) as a way to compute $f(E)$.

We will use a two-dimensional environmental space $E=(E_1, E_2)$ for our discussion and experiment. $E_1$ represents the predefined route clusters while $E_2$ represents the time of the day an AV is deployed. Suppose that we have $m_1$ route clusters, where the routes in the same cluster are considered similar in risk and $m_2$ groups of distinct time of the day. Then the cardinality of the environment space $ |\mathcal{E}|=m_1 \times m_2$. If one were to construct a sequence of $n$ deployments of reasonable size, even if such deployments are evaluated using computer simulation, the number of possible permutations will be $^{m_1\times m_2}P_n $ which is generally a huge number and thus simply intractable.

\subsection{Surrogate Model Form}
To further ensure that $\hat{f}_\theta$ properly track $f$ under well fine-tuned $\theta$, we formulate $\hat{f}_\theta$ as a piecewise linear intensity model \cite{leemis1991nonparametric, zheng2017fitting}. We use the Non-Homogeneous Poisson Process (NHPP) model \cite{burnecki2005modeling, pham2003nhpp} to describe the risk intensities of the traffic system in an area of interest. Although one could presume that the  risky events are positively correlated with traffic densities, i.e. the denser the traffic is, the more likely a risky event would occur, our initial analysis does not support this premise. In Fig \ref{risk_time}, one sees that while the traffic represented by the number of observation during night-time between 00:00 until 04:00 is decreasing overall, the probability of risky events eventually does not, which could be explained by the safety hazard related to nighttime shift driving, fatigue, drowsy driving, etc \cite{vanlaar2008fatigued}.

\begin{figure}
\centerline{
\includegraphics[width=0.95\linewidth]{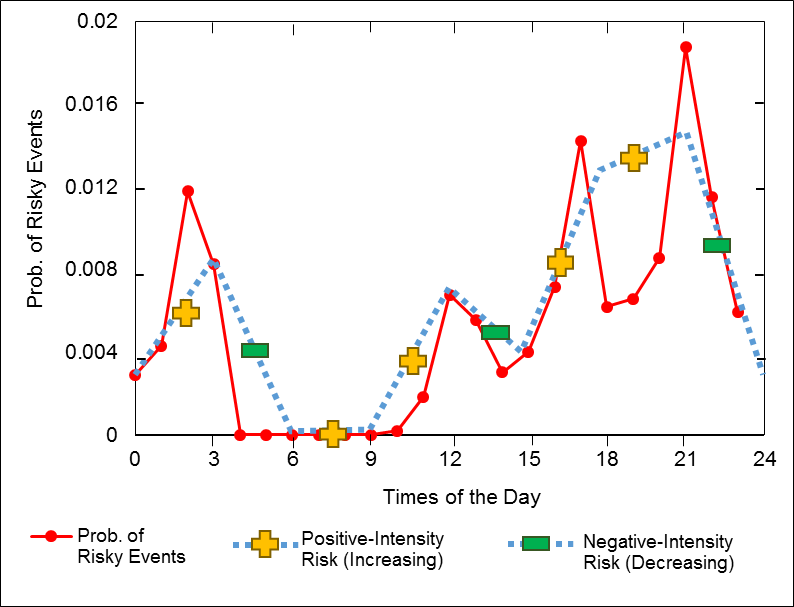}
}
\caption{The use of intensity model in describing the increasing and decreasing driving risk trends in a 24-hour period}
\label{risk_intensity}
\end{figure}

Under NHPP scheme, $\lambda(t)$, which is called intensity function at $t$, represents the rate at which the risk change at time $t$ in the system. It is obvious that such $\lambda(t)$ would be positive for some $t$ before the peak hour period and would be negative after the peak hour period in our context. The term 'piecewise' here refers to the fact that $\lambda(t)$ would be a piecewise function in $t$ consisting of $m_2$ pieces. In Figure \ref{risk_intensity}, we provide a pictorial representation of this notion of risk intensities. With this definition, the cumulative intensity function $\Lambda(t)$, which is defined as
\begin{align}
\Lambda(t) = \int_0^t \lambda(\tau) d\tau,
\label{Lambda_formula}
\end{align}
represents the probability of AV engaging or encountering risky events at time $t$. 

Recall that there are $m_2$ distinct time groups, i.e. the times of the day are grouped into $[0,t_1), [t_1,t_2), \dots, [t_{m_2-1},24)$. To apply (\ref{Lambda_formula}), we should have $\lambda_1, \lambda_2, \dots, \lambda_{m_2}$ and $\lambda(t)=\lambda_k$ when $t$ is in $k$-th time group, for any $t \in [0,24)$.  Therefore, we define $\hat{f}_\theta(E) = \hat{f}_\theta(E_1,E_2)$ as
\begin{align}
\hat{f}_\theta(E_1,E_2) = \theta_1^{l(E_1)} (\theta_{2}^0 + \Lambda(E_2)).
\label{findfhat}
\end{align}
where $l(E_1)$ is the route cluster associated with $E_1$. The parameter vectors can be decomposed into $\theta = [\theta_1, \theta_2]$. Here $\theta_1 = [\theta_1^1,\theta_1^2, \dots, \theta_1^{m_1}]$ and $\theta_2=[\theta_2^0, \theta_2^1, \dots,\theta_2^{m_2-1}]$ are associated with route and time, respectively. In the following analysis, we will refer to $\theta_1$ as the spatial parameter and $\theta_2$ as the temporal parameter for the surrogate model. It shall be obvious that complex capability measure with spatiotemporal pattern can be described with our formulation. From Figure \ref{risk_time} and Figure \ref{risk_space}, we see the temporal factor results in a clear pattern of risk. Furthermore, we observed that some routes are lower in risk while the others are higher. Thus, to reduce the number of parameters, we will later cluster the routes based on the risk and the number of data logged during the experiment. Note that it is sufficient to estimate only the first $m_2-1$ elements of the temporal parameters and let the last intensity $\theta_2^{m_2}$ be dynamically adjusted to model a periodic or cyclic behavior. This notion of periodicity is indeed a feature that we are keen to accommodate to describe the cyclic nature of the traffic densities, with a period of one day. Thus, we could remove $\theta_2^{m_2}$ and append $\theta_2^0$ in $\theta_2$ as an additional parameter to estimate to determine the base level of risk. Therefore, at the end, we have $m_1+m_2$ total number of parameters to estimate under this surrogate model form, which is generally tractable.

\section{Experiment and Findings}
In this section, we present the settings and findings from our numerical experiment.

\begin{figure*}[h]
\centerline{
\includegraphics[width=0.82\linewidth]{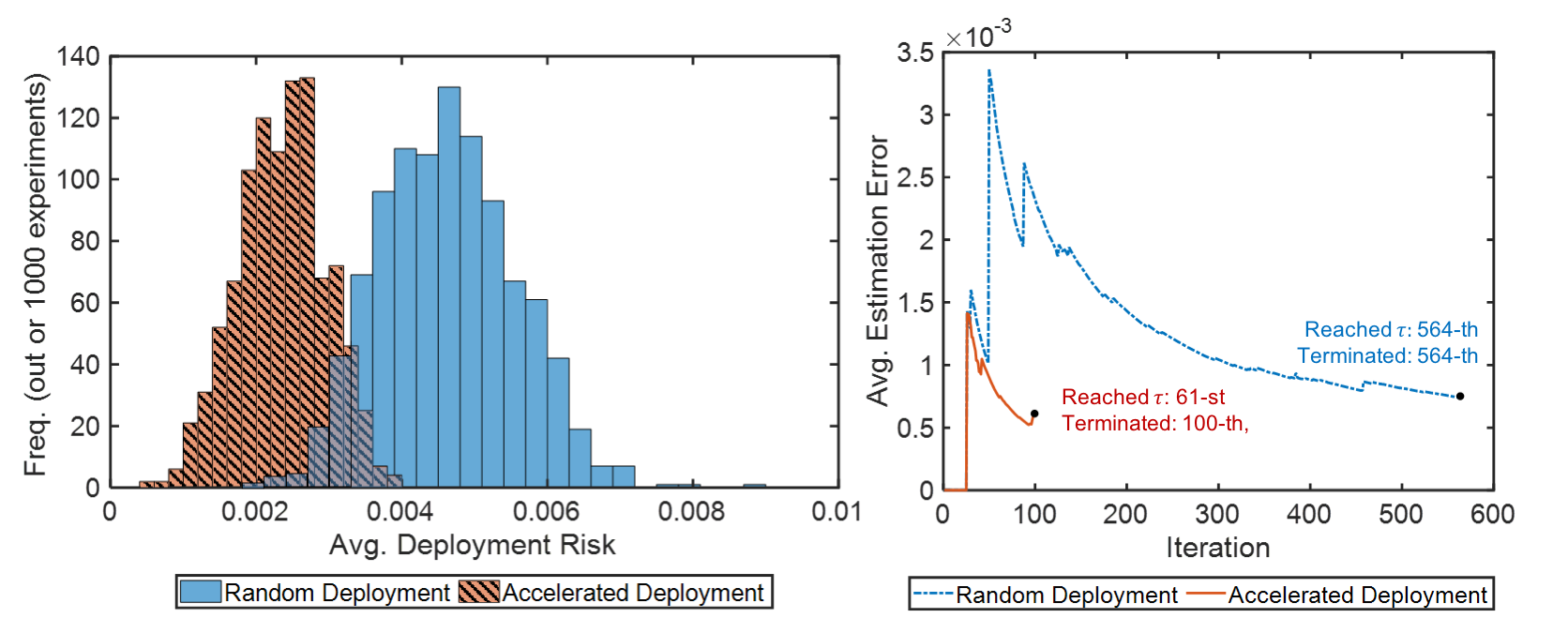}
}
\caption{Numerical performance of the Accelerated Deployment compared to Random Deployment}
\label{performance}
\end{figure*}

\subsection{Experiment}
In our experiment, we used SPMD one-month dataset \cite{spmd} to define the set of deployment environment $\mathcal{E}$ as well as the risk profile $f(E), \forall E \in \mathcal{E}$. We indicated a risky event status for any instances satisfying (i) range $R \leq 10$ feet, (ii) range rate $\dot{R} < 0$. In SPMD terms, range denotes the distance between any logged encounters and the range rate refers to their relative velocity. We conducted a preliminary analysis on the one-month SPMD data, preprocessed, then discretized the environment for tractability reason. In the process, we reduced the number of the routes from 329 as queried from the database by clustering them using $k$-means algorithm into  $m_1=16$ route clusters. Besides, we discretized the 24-hour continuous cyclical time period time into $m_2=8$ time groups, each of which comprise a 3-hour time period.

We then applied the Accelerated Deployment framework to evaluate the capability of the AV system assuming it would be perform at least as good as the driving recorded in the data. As the deployment problem settings, we set $\xi = 0.02$, the tolerance for average estimation error $\tau=7.5^{-4}$ and the minimum number of deployments $\bar{N}=100$. Coping with the number of parameters to estimate, we started the framework with training samples of 25, randomly sampled. The framework terminated with results summarized in Table 1, where we also provide in Figure \ref{performance} a summary of one thousand replicated experiments. Besides, we highlight the estimation accuracy and the search behavior over the number of iterations in Figure \ref{accuracy}.

\begin{table*}
\caption{Numerical performance in terms of risk $f$, estimation error $z$, and number of deployment iterations $n^*$}
\begin{tabularx}{\textwidth}{@{}l*{5}{C}c@{}}
\toprule
Deployment Strategy     & Avg $f$ & Std $f$ & Avg $z$ & $n^*$ & Acc. Ratio\\
\midrule
Accelerated Deployment	& 2.40$\times 10^{-3}$      & 7.30$\times 10^{-3}$        & 5.94$\times 10^{-4}$   & 100    & 5.64 times    \\ 
Random Deployment		& 4.60$\times 10^{-3}$      & 1.12$\times 10^{-2}$        &1.12$\times 10^{-3}$    & 564    & 1.00 (baseline)    \\
\bottomrule
\end{tabularx}
\label{result}
\end{table*}

\subsection{Findings}
Our findings will be discussed in three parts: the overall performance, search behavior, and estimation accuracy trend.

\subsubsection{The overall performance}
From Table \ref{result}, we see that the performance of the Accelerated Deployment framework is superior compared to the random deployment. It is clear that the Accelerated Deployment is less exposed to safety risk compared to random deployment in both mean and standard deviation. While the magnitude may seem too small to notice, we remind the readers that AV deployment problems deal with rare events with safety-critical consequences - a small change in probability of occurrence matters. Furthermore, these risk exposure value is averaged over the number of deployment iterations. With Accelerated Deployment having lower deployment risk, lower spread (in terms of standard deviation), and more importantly, fewer number of iterations, the cumulative deployment risk is significantly less compared to that of random deployment (see the spread and the estimation error trend in Figure \ref{performance}). 

Besides, we also note that we terminate in $n^*=100$ iterations with Accelerated Deployment with an average estimation error of 5.94 $\times 10^{-4}$ and would have terminated in $n^*=564$ with random deployment with $7.48 \times10^{-4}$ average estimation error. Thus, we achieve a 5.64 times faster deployment time, higher accuracy, and more importantly, less risky deployments. It is worth noting that had we ignored the minimum number of $\bar{N}=100$ iterations, the Accelerated Deployment framework would have terminated in only 61 iterations and achieved $7.42 \times 10^{-4}$ average estimation error.

\subsubsection{Search behavior} 
\begin{figure}
\centerline{
\includegraphics[width=0.7\linewidth]{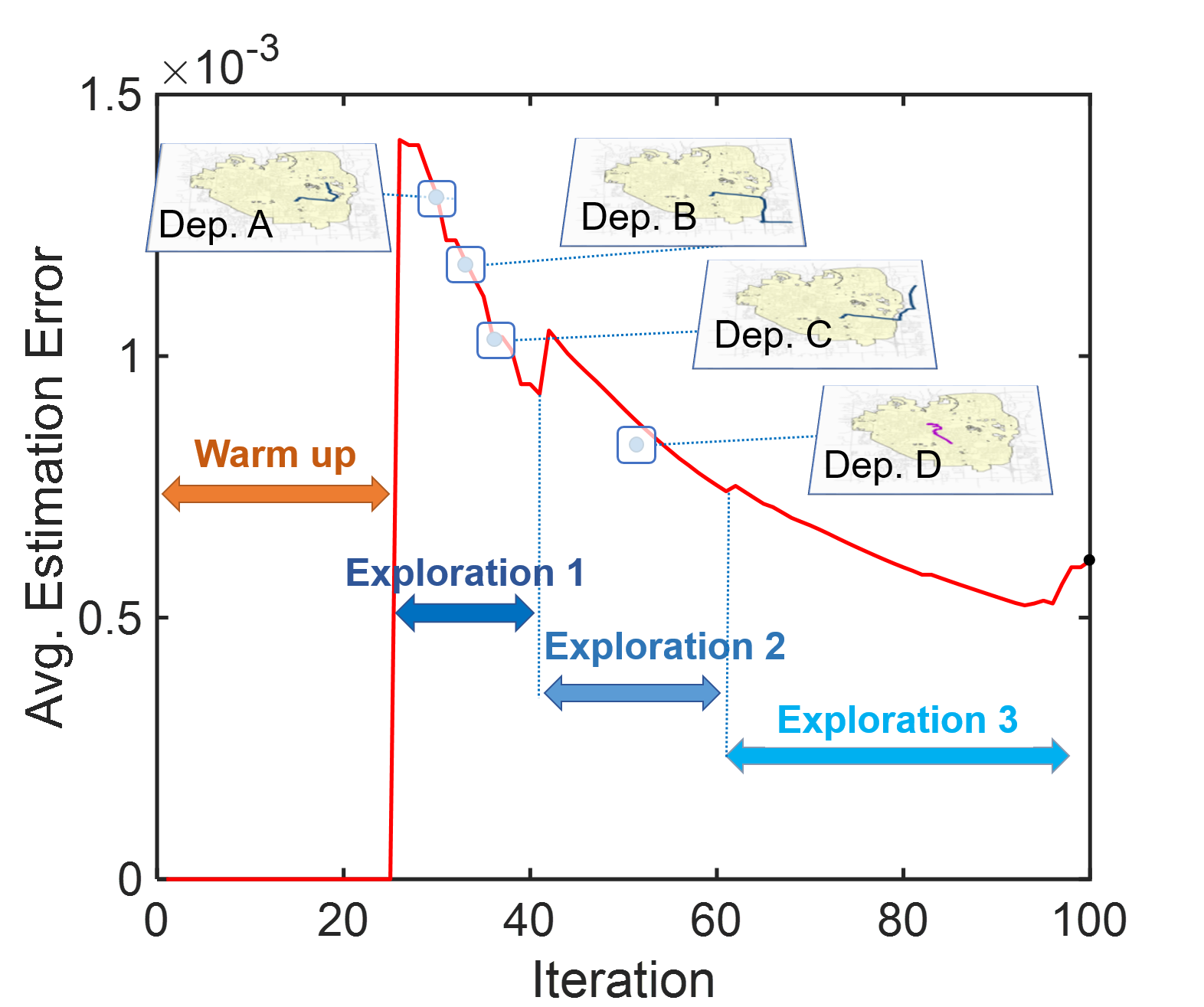}
}
\caption{The estimation error in terms of the number of iterations}
\label{accuracy}
\end{figure}

We investigate how versatile our framework in `managing' the deployment risk. We observe from Figure \ref{risk_space} that some routes can be considered riskier than the other. Our observation from the search behavior highlights that the framework selects deployment on a particular cluster, try its best to obtain good accuracy by deploying in that cluster, then move to the other `unexplored' clusters. In Figure \ref{accuracy}, we show how the earlier iterations (i.e. Deployment A, B, and C) chose Cluster 9 (blue routes) and later (Deployment D) moves to Cluster 14 (purple routes). 

This behavior is also inferred from the sudden increases in the average estimation errors in Figure \ref{accuracy}. When the framework selected an `unxeplored region', the error increases dramatically, suggesting a `new' region exploration is performed.  Realizing this behavior, we note that it is important to ensure that the deployments are not \textit{locally} concentrated around the safe regions. In other words, devising a proper stopping criteria for the deployments that address both exploration degree as well as target accuracy is critical in order to be able to declare a statistical guarantee on the performance of the AV fleet over the whole environment space. Otherwise, one would most likely stop prematurely after guaranteeing the performance level locally \textit{only} on the safe subset of the environment space. 

\section{Conclusion} 
\label{sec:conclusion}
In this study, we propose a safe and efficient framework to test a pre-produced AV on public streets. By adaptively and purposefully selecting the deployment environment after observing the information encoded from all the previous deployments, the proposed Accelerated Deployment framework gradually shifts from risk-averse to risk-taker as the estimation accuracy increases. For this study, we introduced an additional stopping criterion of a minimum number of deployment iterations to deal with this problem at least to some extent. The numerical result shows that the framework is able to achieve a shorter evaluation period, safer deployment operations, and higher accuracy. We also highlight that there might be cases of terminating prematurely based on \textit{local} confidence or over-generalizing the test result, which is a subject of our future research. Put together with the current streams of AV design and evaluation researches, this endeavor would be highly beneficial to significantly increase implementability and statistically guarantee the trustworthiness of the real-world testing for AV systems.

\addtolength{\textheight}{-12cm}  

\bibliographystyle{IEEEtran}
\bibliography{bibs}

% Generated by IEEEtran.bst, version: 1.14 (2015/08/26)
\begin{thebibliography}{10}
\providecommand{\url}[1]{#1}
\csname url@samestyle\endcsname
\providecommand{\newblock}{\relax}
\providecommand{\bibinfo}[2]{#2}
\providecommand{\BIBentrySTDinterwordspacing}{\spaceskip=0pt\relax}
\providecommand{\BIBentryALTinterwordstretchfactor}{4}
\providecommand{\BIBentryALTinterwordspacing}{\spaceskip=\fontdimen2\font plus
\BIBentryALTinterwordstretchfactor\fontdimen3\font minus
  \fontdimen4\font\relax}
\providecommand{\BIBforeignlanguage}[2]{{%
\expandafter\ifx\csname l@#1\endcsname\relax
\typeout{** WARNING: IEEEtran.bst: No hyphenation pattern has been}%
\typeout{** loaded for the language `#1'. Using the pattern for}%
\typeout{** the default language instead.}%
\else
\language=\csname l@#1\endcsname
\fi
#2}}
\providecommand{\BIBdecl}{\relax}
\BIBdecl

\bibitem{fagnant2015preparing}
D.~J. Fagnant and K.~Kockelman, ``Preparing a nation for autonomous vehicles:
  opportunities, barriers and policy recommendations,'' \emph{Transportation
  Research Part A: Policy and Practice}, vol.~77, pp. 167--181, 2015.

\bibitem{wang2017much}
W.~Wang, C.~Liu, and D.~Zhao, ``How much data are enough? a statistical
  approach with case study on longitudinal driving behavior,'' \emph{IEEE
  Transactions on Intelligent Vehicles}, vol.~2, no.~2, pp. 85--98, 2017.

\bibitem{zhao2016accelerated}
D.~Zhao, H.~Peng, S.~Bao, K.~Nobukawa, D.~LeBlanc, and C.~Pan, ``Accelerated
  evaluation of automated vehicles using extracted naturalistic driving data,''
  in \emph{The Dynamics of Vehicles on Roads and Tracks: Proceedings of the
  24th Symposium of the International Association for Vehicle System Dynamics
  (IAVSD 2015), Graz, Austria, 17-21 August 2015}.\hskip 1em plus 0.5em minus
  0.4em\relax CRC Press, 2016, p. 287.

\bibitem{greene2011efficient}
D.~Greene, J.~Liu, J.~Reich, Y.~Hirokawa, A.~Shinagawa, H.~Ito, and T.~Mikami,
  ``An efficient computational architecture for a collision early-warning
  system for vehicles, pedestrians, and bicyclists,'' \emph{IEEE Transactions
  on intelligent transportation systems}, vol.~12, no.~4, pp. 942--953, 2011.

\bibitem{waymo2017}
Waymo, ``{Waymo Safety Report: On the Road to Fully Self-Driving},''
  \url{https://storage.googleapis.com/sdc-prod/v1/safety-report/waymo-safety-report-2017.pdf},
  2017.

\bibitem{kalra2016driving}
N.~Kalra and S.~M. Paddock, ``Driving to safety: How many miles of driving
  would it take to demonstrate autonomous vehicle reliability?''
  \emph{Transportation Research Part A: Policy and Practice}, vol.~94, pp.
  182--193, 2016.

\bibitem{stilgoe2018}
\BIBentryALTinterwordspacing
J.~Stilgoe, ``Machine learning, social learning and the governance of
  self-driving cars,'' \emph{Social Studies of Science}, vol.~48, no.~1, pp.
  25--56, 2018. [Online]. Available:
  \url{https://doi.org/10.1177/0306312717741687}
\BIBentrySTDinterwordspacing

\bibitem{calDMV}
D.~of~Motor Vehicles State~of California, ``{Report of Traffic Collision
  Involving an Autonomous Vehicle (OL 316)},''
  \url{dmv.ca.gov/portal/dmv/detail/vr/autonomous/autonomousveh\_ol316+}, 2018.

\bibitem{huang2017sequential}
Z.~Huang, H.~Lam, and D.~Zhao, ``Sequential experimentation to efficiently test
  automated vehicles,'' in \emph{Simulation Conference (WSC), 2017
  Winter}.\hskip 1em plus 0.5em minus 0.4em\relax IEEE, 2017, pp. 3078--3089.

\bibitem{cox2017design}
V.~Cox, ``Design of experiments,'' in \emph{Translating Statistics to Make
  Decisions}.\hskip 1em plus 0.5em minus 0.4em\relax Springer, 2017, pp. 1--31.

\bibitem{zhao2017accelerated}
D.~Zhao, H.~Lam, H.~Peng, S.~Bao, D.~J. LeBlanc, K.~Nobukawa, and C.~S. Pan,
  ``Accelerated evaluation of automated vehicles safety in lane-change
  scenarios based on importance sampling techniques,'' \emph{IEEE transactions
  on intelligent transportation systems}, vol.~18, no.~3, pp. 595--607, 2017.

\bibitem{glynn1989importance}
P.~W. Glynn and D.~L. Iglehart, ``Importance sampling for stochastic
  simulations,'' \emph{Management Science}, vol.~35, no.~11, pp. 1367--1392,
  1989.

\bibitem{leblanc2013longitudinal}
D.~LeBlanc, S.~Bao, J.~Sayer, and S.~Bogard, ``Longitudinal driving behavior
  with integrated crash-warning system: Evaluation from naturalistic driving
  data,'' \emph{Transportation Research Record: Journal of the Transportation
  Research Board}, no. 2365, pp. 17--21, 2013.

\bibitem{wang2018extracting}
W.~Wang and D.~Zhao, ``Extracting traffic primitives directly from
  naturalistically logged data for self-driving applications,'' \emph{IEEE
  Robotics and Automation Letters}, vol.~3, no.~2, pp. 1223--1229, 2018.

\bibitem{wang2018understanding}
W.~Wang, W.~Zhang, and D.~Zhao, ``Understanding v2v driving scenarios through
  traffic primitives,'' \emph{arXiv preprint arXiv:1807.10422}, 2018.

\bibitem{laugier2011probabilistic}
C.~Laugier, I.~Paromtchik, M.~Perrollaz, Y.~Mao, J.-D. Yoder, C.~Tay,
  K.~Mekhnacha, and A.~N{\`e}gre, ``Probabilistic analysis of dynamic scenes
  and collision risk assessment to improve driving safety,'' \emph{Its
  Journal}, vol.~3, no.~4, pp. 4--19, 2011.

\bibitem{leemis1991nonparametric}
L.~M. Leemis, ``Nonparametric estimation of the cumulative intensity function
  for a nonhomogeneous poisson process,'' \emph{Management Science}, vol.~37,
  no.~7, pp. 886--900, 1991.

\bibitem{zheng2017fitting}
Z.~Zheng and P.~W. Glynn, ``Fitting continuous piecewise linear poisson
  intensities via maximum likelihood and least squares,'' in \emph{Simulation
  Conference (WSC), 2017 Winter}.\hskip 1em plus 0.5em minus 0.4em\relax IEEE,
  2017, pp. 1740--1749.

\bibitem{burnecki2005modeling}
K.~Burnecki and R.~Weron, ``Modeling of the risk process,'' in
  \emph{Statistical Tools for Finance and Insurance}.\hskip 1em plus 0.5em
  minus 0.4em\relax Springer, 2005, pp. 319--339.

\bibitem{pham2003nhpp}
H.~Pham and X.~Zhang, ``Nhpp software reliability and cost models with testing
  coverage,'' \emph{European Journal of Operational Research}, vol. 145, no.~2,
  pp. 443--454, 2003.

\bibitem{vanlaar2008fatigued}
W.~Vanlaar, H.~Simpson, D.~Mayhew, and R.~Robertson, ``Fatigued and drowsy
  driving: A survey of attitudes, opinions and behaviors,'' \emph{Journal of
  Safety Research}, vol.~39, no.~3, pp. 303--309, 2008.

\bibitem{spmd}
USDOT, ``{Safety Pilot Model Deployment – Sample Data, from Ann Arbor,
  Michigan (Documentation)},''
  \url{https://data.transportation.gov/Automobiles/Safety-Pilot-Model-Deployment-Data/a7qq-9vfe},
  2014.

\end{thebibliography}

\end{document}